\documentclass{article}

\usepackage{arxiv}
\usepackage[utf8]{inputenc} 
\usepackage[T1]{fontenc}    
\usepackage{hyperref}       
\usepackage{url}            
\usepackage{booktabs}       
\usepackage{amsfonts}       
\usepackage{nicefrac}       
\usepackage{microtype}      
\usepackage{lipsum}		
\usepackage{graphicx}
\usepackage{natbib}
\usepackage{doi}
\usepackage{amsmath}
\usepackage{multirow}
\usepackage{algorithm}
\usepackage{algorithmic}
\usepackage{amsthm}
\newtheorem{proposition}{Proposition}
\newtheorem{remark}{Remark}

\title{Path-Aligned Constraint Equivariance}

\author{
 Yuhao Wang \\
  National University of Defense Technology\\
  Changsha, China\\
  \texttt{wangyuhao24@nudt.edu.cn} \\
   \And
 Yuanxi Peng \\
  National University of Defense Technology\\
  Changsha, China\\
  \texttt{pyx@nudt.edu.cn} \\
}

\date{}

\hypersetup{
pdftitle={A template for the arxiv style},
pdfsubject={q-bio.NC, q-bio.QM},
pdfauthor={Yuhao Wang},
pdfkeywords={First keyword, Second keyword, More},
}

\begin{document}
\maketitle

\begin{abstract}
    Learned super-resolution of physical fields must satisfy structural constraints---such as divergence-free flow ($\nabla\cdot\mathbf{u}=0$), curl-free fields ($\nabla\times\mathbf{u}=\mathbf{0}$), and pointwise sum-to-one normalization ($\mathbf{1}^\top p=1$)---that standard training objectives do not enforce. 
    Soft penalties trade fidelity against satisfaction via a scalar hyperparameter $\lambda$ but cannot drive the residual to zero; hard-constrained regression eliminates the residual by construction yet converges to the conditional mean, losing small-scale spectral structure. 
    We observe that in Conditional Flow Matching with linear paths, a linear constraint $Ax=\mathbf{c}$ is preserved along the entire transport trajectory whenever the constraint operator is linear, the velocity field is compatible ($Av_\theta=\mathbf{0}$), and the initial noise is consistent ($Ax_0=\mathbf{c}$); on the states a trajectory visits, the latter two conditions are also necessary.
    We term this the path-equivariance principle and instantiate it in PACE (Path-Aligned Constraint Equivariance), a framework that enforces these conditions through a null-space parameterization of the velocity field and a constraint-respecting noise sampler, with no modification to the loss function or the ODE integrator. 
    Across three constraint families, PACE reduces the mean constraint residual by one to five orders of magnitude relative to unconstrained baselines, matching the residual level of the data-generation pipeline itself.
    The proposed framework requires only a closed-form parameterization onto $\ker A$; within this scope, it offers exact constraint enforcement while recovering the spectral content and predictive diversity that conditional-mean regression systematically loses, all under a single unified recipe.
\end{abstract}

\section{Introduction}\label{sec:Introduction}

    Learned super-resolution of physical fields has gradually become a standard acceleration method across scientific computing and visual effects: run a cheap coarse simulation, then use a neural network to restore fine-scale structure. 
    When the output is returned to a physical downstream consumer---a PDE solver, a constraint-satisfying classifier, or a thermodynamically consistent renderer---any violation of the governing structural constraint is compounded and destabilizes the consumer. 
    Three representative constraints concretize this concern: incompressible flow requires $\nabla \cdot \mathbf{u}=0$, conservative force fields require $\nabla \times \mathbf{u}=\mathbf{0}$, and multichannel occupancy fields may require pointwise sum-to-one normalization $\sum_{i}{p_i}=1$. 
    Each can be written as $A\mathbf{x} = \mathbf{c}$ for a linear operator $A$. 
    The central challenge is that standard training objectives make no provision for such constraints. 
    An unconstrained regressor minimizes $\mathbb{E}[\|\mathcal{B}_\theta(x^{\mathrm{LR}}) - x^{\mathrm{HR}}\|^2]$ with no guarantee that $A \mathcal{B}_\theta(x^{\mathrm{LR}}) = \mathbf{c}$; because the objective does not explicitly encode the constraint, optimization provides no general guarantee that the learned outputs will lie on the constraint manifold.
    The natural remedy is a soft penalty $\mathbb{E}[\|\mathcal{B}_\theta(x^{\mathrm{LR}})-x^{\mathrm{HR}}\|^2+\lambda\|A\mathcal{B}_\theta(x^{\mathrm{LR}})-\mathbf{c}\|^2]$, following the PINN template~\citep{raissi2019physicsinformed}.
    This introduces a scalar hyperparameter $\lambda$ that trades reconstruction fidelity against constraint satisfaction. 
    Theoretically, as $\lambda \to \infty$, the squared residual $\|A\mathcal{B}_\theta(x^{\mathrm{LR}})-\mathbf{c}\|^2$ should vanish. 
    In practice, the loss surface exhibits strong anisotropy: the network concentrates violations in small spatial regions, achieving low average penalty $\mathbb{E}[\|A \mathcal{B}_\theta(x^{\mathrm{LR}})-\mathbf{c}\|^2]$ while sharp pointwise violations survive even at large $\lambda$.

    Beyond constraint enforcement, the choice between regression and generation shapes what structure the model recovers. 
    A regression model trained with mean-squared error converges to the conditional expectation $\mathcal{B}_\theta(x^{\mathrm{LR}}) \to \mathbb{E}[x^{\mathrm{HR}} \mid x^{\mathrm{LR}}]$~\citep{bishop2006pattern}. 
    When the conditional distribution $p(x^{\mathrm{HR}} \mid x^{\mathrm{LR}})$ has high-frequency variance, the expectation averages over modes and produces blurred output, losing the small-scale structure that matters for physical fidelity. 
    Generative models offer a principled escape by learning the full conditional distribution and sampling from it. 
    Flow Matching~\citep{Lipman2023FlowMatching, Tong2023OTCFM} parameterizes this distribution through a time-dependent velocity field $v_\theta(x_t, t)$ whose ODE $\dot{x}_t = v_\theta(x_t, t)$ transports a simple noise distribution to the data distribution, yielding sharp, diverse samples that avoid the averaging inherent in regression.
    However, standard Flow Matching provides no mechanism to preserve such constraints along the transport path.

    Our central insight is that a linear physical constraint in Flow Matching is a trajectory property, not an endpoint property.
    In the Conditional Flow Matching framework with linear interpolation paths $x_t = (1-t)x_0 + tx_1$, consider a linear constraint $Ax = \mathbf{c}$ where $A: \mathcal{X} \to \mathcal{Y}$ is a linear operator and $\mathbf{c}$ is a constant target. 
    Assuming the constraint operator $A$ is linear, the constraint holds along the entire trajectory $Ax_t = \mathbf{c}$ for all $t \in [0,1]$ provided two dynamical conditions are met: 
    (1) Velocity-field compatibility: $Av_\theta(x_t, t) = \mathbf{0}$. 
    (2) Endpoint consistency: $Ax_0 = \mathbf{c}$ (the data endpoint $x_1=x^{\mathrm{HR}}$ satisfies $Ax_1=\mathbf{c}$ automatically). 
    These conditions are sufficient in general and, on the states a trajectory visits, also necessary.
    For the three constraint families in this paper, these conditions reduce to concrete architectural choices. 
    For divergence-free flow ($A = \nabla\cdot$, $\mathbf{c}=\mathbf{0}$), the velocity field must satisfy $\nabla\cdot v_\theta = 0$, which is enforced by predicting a stream function $\psi$ and applying $v_\theta = \nabla^\perp \psi$. 
    For curl-free fields ($A = \nabla\times$, $\mathbf{c}=\mathbf{0}$), the velocity field must satisfy $\nabla\times v_\theta = 0$, enforced by predicting a scalar potential $\phi$ and applying $v_\theta = \nabla \phi$. 
    For sum-to-one normalization ($A = \mathbf{1}^\top$, $\mathbf{c}=1$), we work with the affine constraint $\mathbf{1}^\top v_\theta = 0$ (zero-mean deviation), enforced by predicting an unconstrained field $v$ and applying the mean-centering head $v_\theta = v - \frac{1}{C}(\mathbf{1}^\top v)\mathbf{1}$, where the affine sum-to-one membership is recovered at sampling time through a fixed offset.

    This paper makes the following contributions: 
    (1) Formally sufficient conditions for trajectory-level constraint preservation. 
    We prove that for a linear operator, velocity-field compatibility together with endpoint consistency is sufficient for $Ax_t=\mathbf{c}$ at all $t$, and that both conditions are necessary on the states a trajectory visits, giving a rigorous characterization of constraint evolution in generative ODE dynamics. 
    (2) Unified framework across constraint families. Except for the null-space parameterization and noise sampler, we demonstrate that the same architectural recipe applies without modification to three structurally distinct constraint families. This universality suggests that the principle extends beyond the specific operators tested here. 
    (3) Empirical resolution of the fidelity--constraint trade-off. 
    PACE achieves a 26\%--58\% reduction in CRPS over hard regression while reducing constraint residuals to below $10^{-5}$ (differential operators) and $10^{-7}$ (sum-to-one), matching the data-generation noise floor---resolving the fidelity--constraint trade-off that limits unconstrained regressors and soft-penalty baselines.

\section{Related Work}\label{sec:Related Work}

    We organize prior work along a single axis: the strength with which a linear physical constraint is enforced in neural prediction. 
    This ranges from no enforcement, to soft penalties, to exact structural reparameterization in regression, and finally to constraint-preserving generative models. 
    Our contribution sits at the intersection of the last two categories: exact enforcement within a generative framework, applied uniformly across three constraint families.

    \textbf{Unconstrained methods.} 
    Standard super-resolution architectures --- FSRCNN~\citep{Dong2016FSRCNN}, EDSR~\citep{Lim_2017_CVPR_Workshops}, SwinIR~\citep{Liang_2021_ICCV}, U-Net~\citep{Ronneberger2015} --- and operator-learning methods --- FNO~\citep{Li2021}, variants~\citep{Kovachki2023NeuralOperator, Tran2023FactorizedFNO} --- optimize pixelwise fidelity losses without explicit constraint terms. 
    These methods achieve strong perceptual quality on natural images, reflected in high values of fidelity metrics such as PSNR and SSIM, and have been gradually adapted to scientific domains~\citep{Pathak2022FourCastNet, Geneva2022Diffusion}, but make no provision for physical constraints.

    \textbf{Soft constraints.} 
    Physics-Informed Neural Networks~\citep{raissi2019physicsinformed} augment the training loss with a residual penalty $\lambda\|A\mathcal{B}_\theta(x)-\mathbf{c}\|^2$ for the constraint operator $A$. 
    This approach has been instantiated for incompressibility through divergence penalties in turbulent super-resolution~\citep{Gao2021PhysicsInformedSR} and Navier–Stokes solvers~\citep{Jin2021NSsolvers}, and for probability normalization through softmax regularization in segmentation~\citep{Wang2020LongTail, Luo2021DTC}. 
    The hyperparameter $\lambda$ trades data fidelity against constraint satisfaction, and tuning it remains an open challenge: Raissi et al.~\citep{raissi2019physicsinformed} and Thakur et al.~\citep{Thakur2024TemporalConsistency} report that constraint residuals remain sensitive to $\lambda$ and rarely vanish, even at large penalty weights. 
    The fundamental tension is that the constraint is enforced through optimization rather than by construction.
    
    \textbf{Hard constraints in regression.} 
    Several works eliminate constraint residuals by predicting an unconstrained primitive and applying a structural projection. 
    For incompressible flow, Mohan et al.~\citep{Mohan2023HardConstraints} predict a vector potential $\mathbf{\Psi}$ and recover velocity through $\mathbf{u}=\nabla\times\mathbf{\Psi}$;
    the 2D stream-function analogue $\mathbf{u}=\nabla^\perp\psi$ has been used in closure modeling~\citep{Sirignano2023LESClosure}.
    For curl-free fields, the Helmholtz scalar potential $\mathbf{u}=\nabla\phi$ is classical and appears in CNN-based electromagnetism~\citep{Qi2024HDNet}. 
    For sum-to-one normalization, softmax post-processing of unconstrained logits is standard in classification~\citep{Goodfellow2016DeepLearning}. 
    Each of these null-space heads enforces the constraint exactly by construction, but all prior instantiations we are aware of are trained as regressors, which converge to the conditional mean and lose high-frequency structure~\citep{bishop2006pattern}.
    
    \textbf{Generative models with physical constraints.} 
    Recent work has begun to combine generative models with physical constraints. 
    Diffusion-based approaches include physics-informed score matching~\citep{Bastek2024PhysicsInformedDiffusion} and post-step projection in CFD surrogates~\citep{Um2020PostStep}. 
    These methods improve sample diversity compared to regression but enforce constraints either at the sampling endpoint or through iterative correction, without guaranteeing that intermediate states satisfy the constraint. 
    Flow Matching~\citep{Lipman2023FlowMatching} and its optimal-transport variant OT-CFM~\citep{Tong2023OTCFM} parameterize generation as a deterministic ODE $\dot{x}_t = v_\theta(x_t, t)$, offering a natural setting for trajectory-level constraint enforcement. Our work builds on the independent-coupling variant with linear paths, which to our knowledge has not been combined with hard structural heads that preserve $Av_\theta = \mathbf{0}$ by construction.

    \textbf{Gap to our work.} 
    To our knowledge, no prior work enforces a linear physical constraint along the entire transport trajectory of a generative model, nor treats divergence-free, curl-free, and sum-to-one constraints as instances of a single structural recipe. 
    The closest works --- PCFM~\citep{pcfm2025} and PBFM~\citep{Baldan2025PhysicsVD} --- enforce constraints at the endpoint or through post-hoc correction, but do not guarantee $A x_t = \mathbf{c}$ at every integration step. 
    Our framework differs in placing the constraint in the geometry of the velocity field, the initial noise, and the integrator, ensuring path-level preservation by construction.

\section{Methodology}\label{sec:Methodology}

\subsection{Path-Equivariance Principle}\label{sec:Path-Equivariance Principle}
    
    We adopt Conditional Flow Matching with linear paths and independent coupling (CFM), which offers two properties essential for hard constraint preservation. 
    First, the sampling process is a deterministic ODE $\dot{x}_t = v_\theta(x_t, t, x^{\mathrm{LR}})$; any linear operator $A$ commutes with the time derivative $\frac{d}{dt}(Ax_t) = A\dot{x}_t$, enabling trajectory-level constraint enforcement. 
    Second, the velocity target $u_t = x_1 - x_0$ is constant along each linear path $x_t = (1-t)x_0 + tx_1$, allowing pointwise assessment of $v_\theta$. 
    
    Preserving a linear constraint along the entire ODE trajectory confines the integral manifold to the affine constraint set, so intermediate states remain physically consistent without post-hoc correction---turning constraint satisfaction from a sampling-algorithm burden into a natural consequence of the ODE geometry. 
    We therefore seek to guarantee $Ax_t = \mathbf{c}$ for all $t \in [0,1]$ along the linear path 
    $x_t = (1-t)x_0 + tx_1$ under $\dot{x}_t = v_\theta(x_t, t)$. 
    Applying the linear operator $A$ to the ODE and exploiting its linearity to commute with the time derivative yields:
    \begin{equation}
        \frac{d}{dt}(A x_t) \;=\; A \dot{x}_t \;=\; A v_\theta(x_t, t).
    \label{eq:constraint_ode}
    \end{equation}

    \begin{proposition}[Trajectory-level constraint invariance]
    \label{prop:invariance}
    Let $A:\mathcal{X}\to\mathcal{Y}$ be linear and let $x_t$ ($t\in[0,1]$) solve $\dot{x}_t = v_\theta(x_t,t)$ from initial condition $x_0$. 
    If \textbf{(i)} $A x_0 = \mathbf{c}$ and \textbf{(ii)} $A v_\theta(x,t)=\mathbf{0}$, then $A x_t = \mathbf{c}$ for all $t\in[0,1]$.
    \end{proposition}

    \begin{proof}
    By linearity of $A$, it commutes with the time derivative, so 
    $\frac{d}{dt}(A x_t) = A\dot{x}_t = A v_\theta(x_t,t) = \mathbf{0}$ 
    using (ii). Integrating over $[0,t]$ and applying (i),
    \begin{equation}
        A x_t = A x_0 + \int_0^t A v_\theta(x_\tau,\tau)\,d\tau 
              = A x_0 = \mathbf{c}.
    \label{eq:integral}
    \end{equation}
    For the Euler scheme, $A x_{n+1} = A x_n + \Delta t\, A v_\theta(x_n,t_n) 
    = A x_n$; induction from $A x_0 = \mathbf{c}$ gives $A x_n = \mathbf{c}$ 
    for every $n$.
    \end{proof}

    Two points are worth emphasizing. 
    First, Proposition~\ref{prop:invariance} depends only on the linearity of $A$ and the state remaining on the actual ODE trajectory; 
    it does not require the linear interpolation path $x_t=(1-t)x_0+tx_1$, and therefore transfers to any integrator whose steps stay tangent to $\ker A$. 
    Second, conditions (i)--(ii) are also necessary on the reachable set, as we now record.

    \begin{remark}[Partial converse]
    The two conditions are also necessary.
    If $A x_t = \mathbf{c}$ for all $t$, then $A x_0 = \mathbf{c}$ by evaluation at $t=0$, and differentiating the identity gives $A v_\theta(x_t,t) = \mathbf{0}$ at every state the trajectory visits.
    Hence, any construction that preserves the constraint must satisfy conditions~(i) and~(ii) on the reachable set.
    The architectural choice $A v_\theta \equiv \mathbf{0}$ everywhere (Condition~ii) is therefore the minimal state-independent way to guarantee tangency without knowing the trajectory in advance.
    \end{remark}

    Architecturally, conditions (i) and (ii) of Proposition~\ref{prop:invariance} are enforced through two choices applied uniformly across all constraint families below: a null-space head $v_\theta = \Gamma(\mathcal{B}_\theta)$ guaranteeing $A v_\theta = \mathbf{0}$ (Condition~ii), and a constraint-respecting noise sampler $\pi_0$ drawing $x_0$ from $\{x : Ax = \mathbf{c}\}$ (Condition~i). The linearity hypothesis holds by construction for all three operators in Table~\ref{tab:constraints} ($\nabla\cdot$, $\nabla\times$, $\mathbf{1}^\top$).

    \begin{table}[h]
        \centering
        \small
        \setlength{\tabcolsep}{4pt}
        \begin{tabular}{lcccc}
            \toprule
            Constraint  & $C$ & Operator $A$      & Target $\mathbf{c}$   & Null-space primitive           \\
            \midrule
            Div-free    & 2   & $\nabla\cdot$     & $\mathbf{0}$           & stream function $\psi$         \\
            Curl-free   & 2   & $\nabla\times$    & $\mathbf{0}$           & scalar potential $\phi$        \\
            Sum-to-one  & 3   & $\mathbf{1}^\top$ & $1$                    & logits $\ell \in \mathbb{R}^C$ \\
            \bottomrule
        \end{tabular}
        \caption{Three constraint families. $C$: number of data channels; $A$: linear constraint operator; $\mathbf{c}$: target value. The null-space primitive is predicted by the backbone $\mathcal{B}_\theta$ and mapped through the parameterization $\Gamma$ to obtain the velocity field $v_\theta=\Gamma(\mathcal{B}_\theta)\in\ker A$.}
        \label{tab:constraints}
    \end{table}
    
\subsection{Instantiation}\label{sec:Instantiation}

    We enforce the two path-equivariance conditions---endpoint consistency and velocity-field compatibility---by applying the null-space parameterization $\Gamma$ in two places:
    (1) a noise sampler that draws $x_0 \sim \pi_0$ from the constraint set, and
    (2) a network operator that maps the backbone output into $\ker A$ (or into the zero-mean subspace for sum-to-one).
    These are the only modifications beyond standard CFM (Algorithm~\ref{alg:PACE}); both add negligible overhead---one
    FFT--multiply--IFFT pass or pointwise mean subtraction.

    \begin{algorithm}[tb]
    \caption{PACE: Training and Inference}
    \label{alg:PACE}
    \begin{algorithmic}[1]
    \REQUIRE Dataset $\{(x^{\mathrm{LR}}_i,x^{\mathrm{HR}}_i)\}_{i=1}^N$, $\Gamma$, $\pi_0$, $\mathcal{B}_\theta$
    \STATE \textbf{--- Training phase ---}
    \FOR{each iteration}
        \STATE Sample $(x^{\mathrm{LR}},x^{\mathrm{HR}})$ from dataset, $t\sim\mathcal{U}[0,1]$
        \STATE $x_1\gets x^{\mathrm{HR}}$;\; $x_0\sim\pi_0$
        \STATE $x_t\gets(1-t)x_0+tx_1$;\; $u_t\gets x_1-x_0$
        \STATE $v_\theta\gets\Gamma(\mathcal{B}_\theta(x_t,\,t,\,x^{\mathrm{LR}}))$
        \STATE $\theta\gets\theta-\eta\,\nabla_\theta\,\|v_\theta-u_t\|_2^2$
    \ENDFOR
    \REQUIRE $x^{\mathrm{LR}}$, $\Gamma$, $\pi_0$, $\mathcal{B}_\theta$, number of steps $S$
    \STATE \textbf{--- Inference phase ---}
    \STATE $x_0\sim\pi_0$;\; $\Delta t\gets 1/S$
    \FOR{$n=0,\ldots,S-1$}
        \STATE $t_n\gets n\cdot\Delta t$
        \STATE $v_n\gets\Gamma(\mathcal{B}_\theta(x_n,\,t_n,\,x^{\mathrm{LR}}))$
        \STATE $x_{n+1}\gets x_n+\Delta t\cdot v_n$
    \ENDFOR
    \RETURN $x_S$
    \end{algorithmic}
    \end{algorithm}

    \textbf{Noise samplers.}
    For the homogeneous constraints (div-free and curl-free), we draw a scalar Gaussian field $\epsilon \in \mathbb{R}^{B\times 1\times H\times W}$ and set $x_0 = \Gamma(\epsilon)$, which guarantees $Ax_0 = \mathbf{0}$ by construction.
    For the affine sum-to-one constraint $\mathbf{1}^\top x = 1$, we decompose each state as $x = \tilde{x} + \frac{1}{C}\mathbf{1}$, where the deviation $\tilde{x}$ satisfies the homogeneous condition $\mathbf{1}^\top\tilde{x} = 0$.
    Accordingly, we sample $\epsilon \in \mathbb{R}^{B\times 3\times H\times W}$ and set $x_0 = \Gamma_{\mathrm{s}}(\epsilon) + \frac{1}{C}\mathbf{1}$, which satisfies the affine constraint by construction and aligns the sum-to-one case with the homogeneous framework.

    \textbf{Network heads.}
    Let $\mathcal{F}$ denote the 2D FFT, $\hat{g} = \mathcal{F}[g]$, and $\boldsymbol{\kappa} = (\frac{2\pi k_x}{L}, \frac{2\pi k_y}{L})$ the angular wavenumber vector on a domain of size $L=2\pi$. 
    The null-space head $\Gamma$ is applied as a post-processing layer: $v_\theta = \Gamma(\mathcal{B}_\theta(x_t, t, x^{\mathrm{LR}}))$, where $\mathcal{B}_\theta$ outputs a primitive. 
    For divergence-free fields, $\mathcal{B}_\theta$ predicts a stream function $\psi \in \mathbb{R}^{B\times1\times H\times W}$, and $\Gamma_{\mathrm{df}}(\psi) = \mathcal{F}^{-1}\big[\,i\kappa_y\hat{\psi},\; -i\kappa_x\hat{\psi}\,\big] \in \mathbb{R}^{B\times2\times H\times W}$ enforces $\nabla\cdot\Gamma_{\mathrm{df}}(\psi)=0$ by the identity $\nabla\cdot\nabla^\perp=0$. 
    For curl-free fields, $\mathcal{B}_\theta$ predicts a scalar potential $\phi \in \mathbb{R}^{B\times 1\times H\times W}$, and $\Gamma_{\mathrm{cf}}(\phi) = \mathcal{F}^{-1}\big[i\kappa_x\hat{\phi},\, i\kappa_y\hat{\phi}\big] \in \mathbb{R}^{B\times 2\times H\times W}$ enforces $\nabla\times\Gamma_{\mathrm{cf}}(\phi)=0$ by $\nabla\times\nabla=0$. 
    For the sum-to-one constraint, $\mathcal{B}_\theta$ predicts 3 unconstrained channels $\ell \in \mathbb{R}^{B\times 3\times H\times W}$, and $\Gamma_{\mathrm{s}}(\ell) = \ell - \frac{1}{C}(\mathbf{1}^\top\ell)\mathbf{1}$ enforces $\sum_c \Gamma_{\mathrm{s}}(\ell)_c = 0$ pointwise. 
    Since $\Gamma_{\mathrm{s}}$ maps any field to the zero-mean subspace, the velocity satisfies $\mathbf{1}^\top v_\theta = 0$, and the ODE preserves $\mathbf{1}^\top x_t = 1$ for all $t$.

\section{Experimental Setup}

    \textbf{Datasets.} 
    All datasets are defined on the periodic domain $[0,2\pi]^2$ and discretized on a $128\times128$ grid, split into an 8:1:1 ratio for training, validation, and testing. 
    Unless otherwise specified, the main experiments use $4\times$ downsampling for low-resolution inputs, which are generated by uniform downsampling: 
    (1) Divergence-free flow. We utilize a benchmark dataset of two-dimensional Kolmogorov flow~\citep{Baldan2025PhysicsVD} governed by the incompressible Navier–Stokes equations in vorticity–streamfunction formulation. The dataset is simulated on a $128 \times 128$ periodic domain using a pseudo-spectral solver~\citep{liu_2025_15350210} at Reynolds numbers $\mathrm{Re} \in [100,500]$, with a numerical divergence floor of $|\nabla\cdot\mathbf{u}| \approx 6.00\times10^{-6}$. For each $Re$, snapshots are sampled at intervals of $4{,}000 \Delta t$($ \Delta t = 1/\mathrm{Re}$). The dataset contains 40,960 samples.
    (2) Curl-free vector field. 
    Conservative fields are constructed as $\mathbf{u} = \nabla \phi$, where the scalar potential $\phi$ is sampled from a Gaussian random field (GRF) with zero mean and spectral covariance $C_{k_f, \alpha}$. Specifically, $\phi \sim \mathcal{GRF}(0, C_{k_f, \alpha})$ with parameters $k_f = 4$ and $\alpha = 3$, sampled on a $128 \times 128$ periodic grid. The velocity field is then obtained via $\mathbf{u} = \nabla_{\mathrm{spec}} \phi$ using a 2D FFT. By construction, $\nabla \times \mathbf{u} = \mathbf{0}$ to floating-point precision ($\approx 7.87\times10^{-6}$). The dataset contains 10,240 samples.
    (3) Sum-to-one normalization. 
    Per-pixel sum-normalized multichannel fields $p \in \mathbb{R}^3$ (with $\mathbf{1}^\top p = 1$) are generated by applying softmax to three-channel logits $\ell \in \mathbb{R}^{3 \times H \times W}$, where each channel is an independent GRF on a $128\times128$ periodic grid. The sum constraint $\mathbf{1}^\top p = 1$ holds to machine precision ($\approx 1.68 \times 10^{-8}$). The dataset contains 10,240 samples. 

    \textbf{Paradigms.}
    We compare four training configurations across all testbeds to attribute the observed performance gains to each design choice: 
    (1) Unconstrained regression uses a direct MSE loss with no constraint term.
    (2) Soft-penalty regression adds a penalty $\lambda\|Ax-\mathbf{c}\|^2$ (with $\lambda \in \{0.01, 0.1, 1.0\}$).
    (3) Hard regression uses a null-space head $\Gamma$ trained with MSE loss.
    (4) PACE is the full framework, incorporating a null-space head and constraint-respecting noise.
    
    \textbf{Training hyperparameters.}    
    All paradigms use identical backbones, optimizers (AdamW, $\beta_1=0.9$, $\beta_2=0.999$, weight decay $10^{-4}$), learning rate $5\times10^{-4}$ with linear warmup over the first 5 epochs and cosine decay to $5\times10^{-5}$, gradient clipping at $\ell_2$ norm 1.0, and training schedules (300 epochs, batch size 128, EMA decay 0.999). Validation is performed every 5 epochs on a fixed held-out subset, and the best validation-loss checkpoint is used for test-time evaluation.
    
    \textbf{Backbones.}
    The primary backbone is FNO (width 64, depth 4, modes 32, time dim 128). 
    We additionally evaluate U-Net and DiT~\citep{Peebles2023} across all testbeds to verify that the framework is architecture-agnostic.

    \textbf{Evaluation metrics.}
    We evaluate four aspects:
    (1) Pointwise fidelity via PSNR. 
    (2) Constraint satisfaction via the spatial mean $\overline{|r(x)|}$, where $r(x)=Ax-\mathbf{c}$. 
    (3) Spectral quality via the radial energy spectrum $E(k) = \frac{1}{2}\sum_{|\mathbf{k}|=k}|\hat{\mathbf{u}}(\mathbf{k})|^2$ and its absolute deviation $|\Delta E(k)|$. 
    (4) Distribution-level fidelity via the continuous ranked probability score: $CRPS=\frac{1}{CHW}\sum_{c,h,w} \bigl(\mathbb{E}_X|X_{chw}-y_{chw}|-\frac{1}{2}\mathbb{E}_{X,X'}|X_{chw}-X'_{chw}|\bigr)$; the 2-Wasserstein distance $W_2$: mean over $k$ of the 1-D Wasserstein-2 distance between the predicted and true distributions of $E(k)$, and PSNR of the ensemble mean computed over $N$ independent samples.

\section{Results}\label{sec:Results}

\subsection{Constraint satisfaction}\label{sec:Constraint satisfaction}

    \textbf{Main results.}
    Table~\ref{tab:main} reports PSNR and constraint residual across all configurations on the three testbeds.
    PACE reduces $\overline{|r(x)|}$ by one to five orders of magnitude relative to unconstrained regression ($\lambda=0$) and matches the data-generation residual level on all three problems, at a cost of at most ${\sim}2.5$\,dB PSNR. Sweeping $\lambda$ from $10^{-2}$ to $10^{0}$ shows the fundamental limitation of soft penalties: on div-free, the residual plateaus at $2.03{\times}10^{-4}$---over an order of magnitude above the hard methods---and on curl-free it remains at $9.09{\times}10^{-3}$ even at $\lambda{=}1$, indicating that, under the tested optimization settings, increasing the penalty weight does not eliminate the remaining residual.
    PACE and Hard regression coincide on the residual to within floating-point round-off, as both formally restrict outputs to $\ker A$.

    \begin{table*}[t]
    \centering
    \small
    \begin{tabular}{lccccccc}
    \toprule
    Testbed & Metric & $\lambda{=}0$ & $\lambda{=}10^{-2}$ & $\lambda{=}10^{-1}$ & $\lambda{=}10^{0}$ & Hard & PACE \\
    \midrule
    \multirow{2}{*}{Div-free}
            & PSNR & $\mathbf{65.66}$ & 63.05 & 59.66 & 59.25 & 60.64 & 65.65 \\
            & $\overline{|\nabla\cdot\mathbf{u}|}$ & $2.27\times10^{-2}$ & $1.46\times10^{-3}$ & $3.86\times10^{-4}$ & $2.03\times10^{-4}$ & $7.34\times10^{-6}$ & $\mathbf{6.39\times10^{-6}}$ \\
    \midrule
    \multirow{2}{*}{Curl-free}
            & PSNR & 45.37 & 44.03 & 45.22 & $\mathbf{46.24}$ & 42.86 & 44.00 \\
            & $\overline{|\nabla\times\mathbf{u}|}$ & $1.51\times10^{-1}$ & $7.22\times10^{-2}$ & $4.17\times10^{-2}$ & $9.09\times10^{-3}$ & $\mathbf{7.77\times10^{-6}}$ & $8.25\times10^{-6}$ \\
    \midrule
    \multirow{2}{*}{Sum-to-one}
            & PSNR & 57.90 & 57.90 & 57.86 & $\mathbf{58.22}$ & 57.91 & 55.83 \\
            & $\overline{|\mathbf{1}^\top p-1|}$ & $6.97\times10^{-7}$ & $5.82\times10^{-7}$ & $3.05\times10^{-7}$ & $9.91\times10^{-8}$ & $\mathbf{2.39\times10^{-8}}$ & $3.51\times10^{-8}$ \\
    \bottomrule
    \end{tabular}
    \caption{\textbf{Constraint satisfaction and reconstruction fidelity.} 
    PSNR (dB) and mean constraint residual $\overline{|r(x)|}$ across three testbeds ($s=4$). 
    $\lambda=0$ is the unconstrained regressor; $\lambda\in\{10^{-2},10^{-1},10^{0}\}$ are soft-penalty variants. 
    Data-generation residual levels: $6.00{\times}10^{-6}$ (div-free), $7.87{\times}10^{-6}$ (curl-free), $1.68{\times}10^{-8}$ (sum-to-one).}
    \label{tab:main}
    \end{table*}

    On the pointwise sum-to-one constraint, the soft penalty at $\lambda{=}1$ approaches the noise floor, while on both differential operators---div-free and curl-free---it plateaus orders of magnitude above: the tested soft penalties are substantially less effective for the differential constraints than for the pointwise sum-to-one constraint.
    PACE eliminates this operator-dependence entirely: the null-space head $\Gamma$ embeds the constraint into the velocity field geometry, reducing all three families to their respective data-generation residual levels with a single architectural recipe.

\subsection{Generative fidelity}\label{sec:Generative fidelity}

    \textbf{Spectral recovery.}
    Hard regression and PACE coincide on the constraint residual (Table~\ref{tab:main}) but differ fundamentally in how they allocate physical energy across frequencies---a difference that follows directly from the path-equivariance conditions.
    Regression minimizes MSE and therefore estimates the conditional mean, which systematically attenuates variance and smooths small-scale structure~\citep{bishop2006pattern}.
    PACE, by contrast, draws samples from the learned velocity field, producing an ensemble that approximates the conditional distribution.
    Crucially, the null-space head $v_\theta = \Gamma(\mathcal{B}_\theta)$ ensures that the velocity field lies in $\ker A$ by construction: every Euler step advances the state strictly within the constraint manifold, so no spectral energy is diverted into constraint-violating modes.
    We quantify the difference by decomposing the radially averaged energy spectrum into a low band $k < k_c$ and a high band $k \ge k_c$, with $k_c = 32$ the FNO mode cut-off, and reporting $|\Delta E(k)| = |E_{\text{pred}}(k) - E_{\text{true}}(k)|$ in each band (Table~\ref{tab:spec}).

    \begin{table}[!htbp]
        \centering
        \small
        \begin{tabular}{lccc}
            \toprule
            Method & Div-free & Curl-free & Sum-to-one \\
            \midrule
            Hard low-$k$  & $4.67\times10^{-4}$ & $1.44\times10^{0}$  & $6.39\times10^{0}$  \\
            Hard high-$k$ & $4.52\times10^{-7}$ & $5.26\times10^{-6}$ & $7.77\times10^{-5}$ \\
            PACE low-$k$  & $3.01\times10^{-4}$ & $1.47\times10^{-1}$ & $4.93\times10^{-4}$ \\
            PACE high-$k$ & $7.27\times10^{-8}$ & $5.13\times10^{-6}$ & $4.18\times10^{-6}$ \\
            \bottomrule
        \end{tabular}
        \caption{\textbf{Spectral fidelity gap.} $\left|\Delta E(k)\right|$ averaged in each band. Lower value indicates closer proximity to the true spectrum. Here, low-$k$ refers to $k < k_c$ and high-$k$ refers to $k \ge k_c$, with the band split at the FNO mode cut-off $k_c\!=\!32$.}
        \label{tab:spec}
    \end{table}

    PACE yields a smaller spectral deviation than hard regression in both frequency bands on all three testbeds.
    PACE substantially reduces the spectral gap relative to Hard regression: on sum-to-one, the low-band deviation shrinks from $6.39$ to $4.93{\times}10^{-4}$ (four orders of magnitude); on curl-free, from $1.44$ to $0.147$ (${\sim}10{\times}$); on div-free, PACE maintains offsets at $10^{-4}$ and $10^{-8}$ in both bands, respectively.
    The mechanism is that $Av_\theta = \mathbf{0}$ prevents the velocity field from leaking energy into the null-space complement of $A$: in the spectral null-space heads, $\Gamma_{\mathrm{df}}$ and $\Gamma_{\mathrm{cf}}$ restrict each Fourier mode to the subspace orthogonal (div-free) or parallel (curl-free) to its wavevector $\boldsymbol{\kappa}$, so no energy can enter the constraint-violating directional component at any wavenumber.

    \textbf{Distribution-level evidence.}
    The spectral analysis establishes that individual PACE samples recover the physical energy spectrum more faithfully than the conditional mean. 
    We now evaluate whether the generative model provides more accurate probabilistic predictions than the deterministic conditional mean. 
    Table~\ref{tab:distribution} reports ensemble distribution metrics.

    \begin{table}[!htbp]\centering\small
    \setlength{\tabcolsep}{4pt}
        \begin{tabular}{lcccc}
            \toprule
            Testbed & Method & PSNR & CRPS & Spectral $W_2$ \\
            \midrule
            \multirow{2}{*}{Curl-free}
                    & Hard & 42.86 & $5.02\times10^{-2}$ & 4.27 \\
                    & PACE & 43.96 & $2.71\times10^{-2}$ & 1.04 \\
            \midrule
            \multirow{2}{*}{Div-free}
                    & Hard & 60.65 & $1.04\times10^{-3}$ & $3.22\times10^{-3}$ \\
                    & PACE & 63.18 & $4.35\times10^{-4}$ & $3.18\times10^{-3}$ \\
            \midrule
            \multirow{2}{*}{Sum-to-one}
                    & Hard & 57.83 & $9.61\times10^{-4}$ & $7.22\times10^{-4}$ \\
                    & PACE & 57.79 & $7.11\times10^{-4}$ & $1.47\times10^{-3}$ \\
            \bottomrule
        \end{tabular}
        \caption{\textbf{Distribution-level ensemble metrics.} PACE ($N=10$ ensemble) vs.\ Hard regression across three testbeds ($s=4$, FNO backbone). Lower CRPS indicates better probabilistic forecast than deterministic MAE.}
        \label{tab:distribution}
    \end{table}

    Across all three testbeds, PACE achieves a strictly lower CRPS than the Hard regression MAE, with relative improvements of 26\%--58\%.
    Since CRPS is a proper scoring rule that collapses to MAE for deterministic forecasts~\citep{gneiting2007strictly}, a lower value means the entire ensemble distribution is closer to the observation in expectation; the gap (CRPS $<$ MAE) implies that the stochastic ensemble provides a more accurate probabilistic forecast than the deterministic conditional mean. 
    The ensemble-mean PSNR further corroborates this: on curl-free it rises from $42.86$\,dB (Hard) to $43.96$\,dB, and on div-free from $60.65$\,dB to $63.18$\,dB---the classic signature of ensemble averaging reducing uncorrelated per-member error.
    The sum-to-one testbed shows negligible PSNR change ($-0.04$\,dB), which we attribute to the low-information nature of the softmax-GRF prior. 
    Spectral $W_2$ provides complementary frequency-domain evidence. 
    On curl-free, PACE reduces $W_2$ from $4.27$ to $1.04$ ($4.1{\times}$ improvement), confirming that the ensemble spectral distribution better matches the true spectrum. 
    On div-free, the two methods yield nearly identical $W_2$, consistent with the fact that the hard-regression predictor already captures the Kolmogorov energy spectrum well. 
    The sum-to-one result reverses, which may partly reflect sampling variability in high-wavenumber bins at the relatively small ensemble size $N=10$.

    \begin{figure}[!htbp]
        \includegraphics[width=\linewidth]{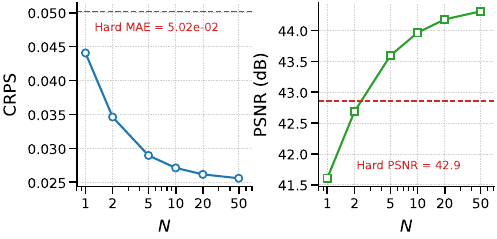}
        \caption{\textbf{Ensemble-size sensitivity on the curl-free testbed.}
        CRPS and ensemble-mean PSNR as functions of ensemble size $N$ ($s{=}4$, FNO backbone).}
        \label{fig:ensemble_sweep}
    \end{figure}

    Figure~\ref{fig:ensemble_sweep} confirms that these gains are not artifacts of a particular ensemble size: on the curl-free testbed, CRPS decreases monotonically from $4.41{\times}10^{-2}$ ($N{=}1$) to $2.56{\times}10^{-2}$ ($N{=}50$), a $42\%$ reduction, while ensemble-mean PSNR rises from $41.61$\,dB to $44.31$\,dB. 
    The rate of improvement decays rapidly beyond $N{=}10$, justifying the default ensemble size used throughout this paper. 

    \subsection{Ablation}
    \begin{table}[!htbp]
        \centering
        \small
        \begin{tabular}{lccc}
            \toprule
            Scale factor & Div-free & Curl-free & Sum-to-one \\
            \midrule
            $s=4$ & $6.39\times10^{-6}$ & $8.25\times10^{-6}$ & $3.51\times10^{-8}$ \\
            $s=8$ & $6.39\times10^{-6}$ & $8.21\times10^{-6}$ & $3.52\times10^{-8}$ \\
            \bottomrule
        \end{tabular}
        \caption{\textbf{Scale ablation.} FNO + PACE at $s=4$ vs $s=8$ on all three testbeds. Constraint residual remains effectively unchanged across scale factors.}
        \label{tab:scale_ablation}
    \end{table}

    \textbf{Scale and backbone robustness.} 
    We verify that the constraint satisfaction properties of PACE are not artifacts of a particular resolution or architecture. 
    At $s=8$, the constraint residual remains at the data-generation residual level on all three testbeds (Table~\ref{tab:scale_ablation}). 

    \begin{table}[!htbp]\centering\small
    \begin{tabular}{llcc}
        \toprule
        Testbed & Model & PSNR & \textbf{$\overline{|r(x)|}$} \\
        \midrule
        \multirow{3}{*}{Div-free}
                &U-Net & 60.47 & $6.39\times10^{-6}$ \\
                &DiT   & 63.89 & $6.39\times10^{-6}$ \\
                &FNO   & 65.65 & $6.39\times10^{-6}$ \\
        \midrule
        \multirow{3}{*}{Curl-free}
                &U-Net & 40.10 & $9.14\times10^{-6}$ \\
                &DiT   & 38.01 & $7.65\times10^{-6}$ \\
                &FNO   & 44.00 & $8.25\times10^{-6}$ \\
        \midrule
        \multirow{3}{*}{Sum-to-one}
                &U-Net & 56.67 & $3.51\times10^{-8}$ \\
                &DiT   & 47.33 & $3.51\times10^{-8}$ \\
                &FNO   & 55.83 & $3.51\times10^{-8}$ \\
        \bottomrule
    \end{tabular}
    \caption{\textbf{Backbone ablation under PACE.} Constraint residual stays at the data-generation residual level for every architecture on every testbed ($s=4$).}
    \label{tab:backbone}
    \end{table}    
    Replacing FNO with U-Net (CNN) and DiT (Transformer) confirms that the residual floor holds independent of the backbone (Table~\ref{tab:backbone}): across all nine architecture--testbed combinations, $\overline{|r(x)|}$ remains at the data-generation residual level, establishing that constraint validity is a property of the null-space parameterization $\Gamma$ and noise sampler $\pi_0$, not of any particular network design.

    \begin{table}[!htbp]\centering\small
        \begin{tabular}{llcc}
            \toprule
            Testbed & Method & PSNR & \textbf{$\overline{|r(x)|}$} \\
            \midrule
            \multirow{8}{*}{Curl-free}
                &Unconstrained FM   & 44.73 & $1.38\times 10^{-1}$ \\
                &Velocity-Only      & 25.30 & $2.92\times 10^{1}$ \\
                &Noise-Only         & 24.49 & $1.12\times 10^{1}$ \\
                &Endpoint-Only      & 43.92 & $6.32\times 10^{-6}$ \\
                &Per-Step           & 41.44 & $6.44\times 10^{-6}$ \\
                &PBFM               & 44.43 & $2.14\times 10^{-1}$ \\
                &PCFM               & 43.02 & $3.10\times 10^{-1}$ \\
                &PACE (Ours)        & 44.00 & $8.25\times 10^{-6}$ \\
            \midrule
            \multirow{8}{*}{Div-free}
                &Unconstrained FM   & 65.31 & $5.03\times 10^{-2}$ \\
                &Velocity-Only      & 25.16 & $2.68\times 10^{1}$ \\
                &Noise-Only         & 56.54 & $2.91\times 10^{-1}$ \\
                &Endpoint-Only      & 63.14 & $1.10\times 10^{-6}$ \\
                &Per-Step           & 53.59 & $1.11\times 10^{-6}$ \\
                &PBFM               & 55.02 & $4.88\times 10^{-2}$ \\
                &PCFM               & 63.38 & $2.08\times 10^{-2}$ \\
                &PACE (Ours)        & 65.65 & $6.39\times 10^{-6}$ \\
            \bottomrule
        \end{tabular}
        \caption{\textbf{Component ablation.} Reconstruction fidelity and constraint satisfaction across eight configurations on the curl-free and div-free testbeds.}
        \label{tab:ablation_endpoint}
    \end{table}

    \textbf{Component ablation.}
    Table~\ref{tab:ablation_endpoint} compares PACE against component-wise ablations and external constrained-generation baselines across both the curl-free and divergence-free testbeds.
    We use the orthogonal Helmholtz projections $\Pi_{\mathrm{cf}}$ and $\Pi_{\mathrm{df}}$, defined in Fourier space as $\widehat{\Pi_{\mathrm{cf}} v}(\boldsymbol{\kappa}) = \boldsymbol{\kappa}(\boldsymbol{\kappa}^\top\hat v)/\|\boldsymbol{\kappa}\|^2$ and $\widehat{\Pi_{\mathrm{df}} v}(\boldsymbol{\kappa}) = \widehat{v}(\boldsymbol{\kappa}) - \boldsymbol{\kappa}(\boldsymbol{\kappa}^\top\hat v)/\|\boldsymbol{\kappa}\|^2$, which map an arbitrary vector field to its curl-free and divergence-free components respectively. 
    Velocity-Only (ablating the noise prior) collapses on both testbeds (curl-free PSNR $25.30$, curl $2.92\times10^{1}$; div-free PSNR $25.16$, divergence $2.68\times10^{1}$), while Noise-Only (ablating the null-space head) retains modest PSNR on div-free ($56.54$\,dB) but fails to suppress divergence below $2.91\times10^{-1}$, and its curl reaches $1.12\times10^{1}$ on the curl-free testbed---four to five orders of magnitude above the PACE residual floor.
    Both $Av_\theta = \mathbf{0}$ and $Ax_0 = \mathbf{c}$ are therefore individually necessary, and only their combination reaches PSNR parity with machine-precision constraint satisfaction. 
    
    Unconstrained FM provides an unconstrained reference for reconstruction fidelity (curl-free PSNR $44.73$\,dB, div-free PSNR $65.31$\,dB), but exhibits severe constraint violation (curl $\overline{|\nabla\times\mathbf{u}|}=1.38\times10^{-1}$, divergence $\overline{|\nabla\cdot\mathbf{u}|}=5.03\times10^{-2}$), confirming that standard Flow Matching learns no awareness of either manifold. 
    PBFM applies an endpoint constraint penalty at $t=1$ via Pareto ConFIG~\citep{Liu2024ConFIG}, yielding curl degradation ($2.14\times10^{-1}$ at PSNR $44.43$\,dB) and modest divergence improvement ($4.88\times10^{-2}$ at PSNR $55.02$\,dB), indicating that the tested endpoint-level soft constraint does not reliably steer the learned vector field toward the constrained set. 
    Endpoint-Only applies the Helmholtz projection once at $t=1$, reducing curl to $6.32\times10^{-6}$ (PSNR $43.92$) and divergence to $1.10\times10^{-6}$ (PSNR $63.14$), with PSNR penalties of $0.81$\,dB and $2.17$\,dB. 
    Per-Step applies $\Pi$ after every Euler advance, maintaining machine-precision residuals (curl $6.44\times10^{-6}$, divergence $1.11\times10^{-6}$) but incurring larger tax ($3.29$\,dB on curl-free, $11.72$\,dB on div-free) possibly because repeated projections move the state away from the distribution encountered during unconstrained model training, causing errors to accumulate.
    PCFM applies per-step Newton projection to enforce constraints during sampling. On curl-free, PCFM yields PSNR $43.02$\,dB with curl $3.10\times10^{-1}$, worse than Unconstrained FM. On div-free, PCFM achieves a $2.4\times$ divergence reduction ($2.08\times10^{-2}$) at $1.93$\,dB cost, consistent with its design for aggregate constraints, yet the residual gap to architectural baselines confirms that per-step Newton correction cannot substitute for training-time constraint embedding.
    PACE provides the strongest overall fidelity–constraint trade-off among the evaluated methods: curl-free PSNR $44.00$\,dB (within $0.73$\,dB of unconstrained) with curl $8.25\times10^{-6}$; div-free PSNR $65.65$\,dB with divergence $6.39\times10^{-6}$. 
    Among the evaluated methods, only PACE remains within $1$\,dB of the unconstrained reference while reaching the numerical constraint floor on both testbeds, confirming that architectural hard constraints eliminate the trade-off between reconstruction fidelity and physical consistency.

    \textbf{Trajectory-level verification.}
    The endpoint metrics above cannot distinguish per-step post-hoc correction from architectural constraint satisfaction.
    Table~\ref{tab:ablation_trajectory_combined} resolves this ambiguity by tracing the constraint residual $\overline{|r(x_t)|}$ along the ODE integration path on both testbeds. 
    Endpoint-Only exhibits monotonic intrinsic drift on both testbeds: although the residual decreases along the trajectory (curl from $29.1$ to $0.263$; divergence from $5.21$ to $2.17\times10^{-2}$), it never reaches the data floor, because the learned vector field drives samples toward the constraint manifold but offers no guarantee of reaching it---Endpoint-Only is approximately identical to the unconstrained trajectory at all $t<1$. 
    Per-Step corrects drift immediately: at $t=0.2$, the residual collapses by seven orders of magnitude (curl from $29.1$ to $3.42\times10^{-6}$; divergence from $5.21$ to $6.13\times10^{-7}$), confirming that the Helmholtz projection is an effective trajectory-level corrector. 
    However, the accumulated PSNR penalty ($3.29$\,dB on curl-free, $11.72$\,dB on div-free) reveals that this correction is reactive: each projection rectifies the current state but cannot prevent the model from generating constraint-violating velocity estimates at the next step. 
    PACE achieves continuous invariance on both testbeds: the residual is maintained at or below the data-generation floor uniformly across all $t\in[0,1]$, including $t=0$ where structured noise initialization is used, demonstrating that the constraint manifold is an invariant manifold of the PACE dynamical system---once initialized on the manifold, the trajectory never leaves it.

    \begin{table}[!htbp]
    \centering\small
    \setlength{\tabcolsep}{4pt}
        \begin{tabular}{llccc}
            \toprule
            Testbed & $\mathbf{t}$ & {Endpoint-Only} & {Per-Step} & {PACE} \\
            \midrule
            \multirow{6}{*}{Curl-free}
            &$0.0$ & $2.91\times 10^{1}$ & $2.91\times 10^{1}$ & $5.48\times 10^{-6}$ \\
            &$0.2$ & $2.26\times 10^{1}$ & $3.42\times 10^{-6}$ & $5.45\times 10^{-6}$ \\
            &$0.4$ & $1.62\times 10^{1}$ & $3.45\times 10^{-6}$ & $5.79\times 10^{-6}$ \\
            &$0.6$ & $1.30\times 10^{1}$ & $4.10\times 10^{-6}$ & $6.06\times 10^{-6}$ \\
            &$0.8$ & $6.56\times 10^{0}$ & $5.16\times 10^{-6}$ & $6.78\times 10^{-6}$ \\
            &$1.0$ & $2.63\times 10^{-1}$ & $6.43\times 10^{-6}$ & $7.75\times 10^{-6}$ \\
            \midrule
            \multirow{6}{*}{Div-free}
            &$0.0$ & $5.21\times 10^{0}$  & $5.21\times 10^{0}$   & $9.13\times 10^{-7}$ \\
            &$0.2$ & $4.05\times 10^{0}$  & $6.13\times 10^{-7}$  & $8.33\times 10^{-7}$ \\
            &$0.4$ & $2.88\times 10^{0}$  & $6.10\times 10^{-7}$  & $8.35\times 10^{-7}$ \\
            &$0.6$ & $2.30\times 10^{0}$  & $7.14\times 10^{-7}$  & $8.68\times 10^{-7}$ \\
            &$0.8$ & $1.13\times 10^{0}$  & $8.91\times 10^{-7}$  & $9.92\times 10^{-7}$ \\
            &$1.0$ & $2.17\times 10^{-2}$ & $1.10\times 10^{-6}$  & $1.17\times 10^{-6}$ \\
            \bottomrule
        \end{tabular}
        \caption{Mean constraint residual $\overline{|r(x_t)|}$ along the ODE trajectory at integration times $t\in\{0.0,0.2,0.4,0.6,0.8,1.0\}$, where $A=\nabla\times$ for the curl-free testbed and $A=\nabla\cdot$ for the divergence-free testbed. For Endpoint-Only, the value at $t=1$ is measured immediately before the final endpoint projection; its post-projection residual is reported in Table~\ref{tab:ablation_endpoint}.}
        \label{tab:ablation_trajectory_combined}
    \end{table}

    Taken together, these results establish a hierarchy of constraint enforcement mechanisms. 
    Velocity-Only, which retains the null-space velocity head but removes the structured noise prior, starts outside the constraint set and therefore cannot recover pathwise invariance. Noise-Only, which retains constraint-respecting initial noise but removes the null-space head, accumulates constraint drift during ODE integration. 
    Soft endpoint penalties provide only limited residual reduction, whereas endpoint and per-step projections can reduce the sampled-state residual but introduce additional reconstruction error. 
    In contrast, PACE embeds the constraint in both the initial distribution and the velocity-field geometry, thereby preserving the constraint throughout the trajectory while maintaining competitive generative fidelity.  
    The trajectory measurements in Table~\ref{tab:ablation_trajectory_combined} further support this interpretation: per-step projection corrects violations after they occur, whereas PACE prevents such violations by keeping the trajectory within the constraint set from initialization onward, up to numerical discretization and floating-point error.

\section{Conclusion}

    We presented the path-equivariance principle, a geometric condition under which flow-based generative models preserve linear constraints to machine precision, and instantiated it in PACE, a training framework that modifies flow matching with a null-space head $\Gamma$ and a constraint-aligned noise prior. 
    Across three constraint families and three backbone architectures, PACE reduces constraint residuals by one to five orders of magnitude relative to unconstrained baselines, matching the residual level of the data-generation pipeline, while recovering spectral content that conditional-mean regression systematically smooths away.
    PACE achieves a 26\%--58\% reduction in CRPS relative to Hard regression across all three testbeds, confirming that the generative ensemble provides more accurate probabilistic forecasts than the deterministic conditional mean without sacrificing constraint satisfaction. 
    Within its applicable scope---linear constraints admitting a closed-form null-space head---the framework offers a rigorous alternative to soft penalties, demonstrating that exact constraint satisfaction and high-fidelity generation are complementary, not competing, objectives. 
    Extending the principle to nonlinear constraints remains open for future investigation.

\bibliographystyle{unsrt}
\bibliography{references}

\end{document}